\documentclass[lettersize,journal]{IEEEtran}
\IEEEoverridecommandlockouts
\pdfoutput=1
\usepackage{cite}
\usepackage{amsmath,amssymb,amsfonts}
\usepackage{algorithmic}
\usepackage{graphicx}
\usepackage{textcomp}
\usepackage{xcolor}
\usepackage{custom-macros}
\usepackage{paralist}
\usepackage{float}
\usepackage{multirow}
\usepackage{booktabs}
\usepackage{amsmath}
\usepackage{soul}
\usepackage{enumitem}
\usepackage{pifont}
\def\BibTeX{{\rm B\kern-.05em{\sc i\kern-.025em b}\kern-.08em
    T\kern-.1667em\lower.7ex\hbox{E}\kern-.125emX}}
\usepackage{xparse}
\usepackage{gnuplottex}
\usepackage{tikz}
\ExplSyntaxOn
\DeclareExpandableDocumentCommand{\convertlen}{ O{cm} m }
 {
  \dim_to_decimal_in_unit:nn { #2 } { 1 #1 } cm
 }
\ExplSyntaxOff

\begin{document}
\bstctlcite{IEEEexample:BSTcontrol}
\title{Energy-efficient DNN Inference on Approximate Accelerators Through Formal Property Exploration}

\author{\IEEEauthorblockN{Ourania Spantidi,
Georgios Zervakis,
Iraklis Anagnostopoulos,~\IEEEmembership{Member,~IEEE},
and J{\"o}rg Henkel,~\IEEEmembership{Fellow,~IEEE}
}

\vspace{-15pt}

\thanks{Ourania Spantidi and Iraklis Anagnostopoulos are with the School of Electrical, Computer and Biomedical Engineering, at Southern Illinois University, Carbondale 62901, USA.}
\thanks{Georgios Zervakis and J{\"o}rg Henkel are with Chair of Embedded System (CES) Department of Computer Science at Karlsruhe Institute of Technology (KIT), Karlsruhe 76131, Germany.}
\thanks{Manuscript received April 07, 2022; revised June 11, 2022; accepted July 05, 2022. This article was presented in the International Conference on Compilers, Architectures, and Synthesis for Embedded Systems (CASES) 2022 and appears as part of the ESWEEK-TCAD special issue.\\
Corresponding author: Ourania Spantidi (ourania.spantidi@siu.edu)}
\thanks{
This work is supported in parts by the German Research Foundation (DFG) through the project “ACCROSS: Approximate Computing aCROss the System Stack” HE 2343/16-1 and 
grant NSF IIP 1361847 from the NSF I/UCRC for Embedded Systems at SIUC.}
}

\maketitle

\begin{abstract}
Deep Neural Networks (DNNs) are being heavily utilized in modern applications and are putting energy-constraint devices to the test. To bypass high energy consumption issues, approximate computing has been employed in DNN accelerators to balance out the accuracy-energy reduction trade-off. However, the approximation-induced accuracy loss can be very high and drastically degrade the performance of the DNN. 
Therefore, there is a need for a fine-grain mechanism that would assign specific DNN operations to approximation in order to maintain acceptable DNN accuracy, while also achieving low energy consumption. 
In this paper, we present an automated framework for weight-to-approximation mapping enabling formal property exploration for approximate DNN accelerators. At the MAC unit level, our experimental evaluation surpassed already energy-efficient mappings by more than $\times2$ in terms of energy gains, while also supporting significantly more fine-grain control over the introduced approximation.
\end{abstract}


\section{Introduction}

Deep Neural Networks (DNNs) are consistently pushing the computing limitations of modern embedded devices. Current applications impose strict accuracy requirements making DNNs essential components in multiple continuously advancing domains~\cite{jouppi2017datacenter},  resulting also in deeper and more complex implementations. The computing requirements of state-of-art DNNs have increased so much that a single inference requires billions of multiply and accumulate (MAC) operations. As embedded devices are resource-constrained (i.e., limited computing and power capabilities), they integrate hardware accelerators, which comprise large amounts of MAC units, to balance out the accuracy/throughput requirements, e.g., 4K MAC units in the Google Edge TPU~\cite{cass2019taking} and 6K MAC units in the Samsung embedded NPU~\cite{park20219}).
However, such a high number of MAC units operating in parallel and performing billions of operations per second results in elevated energy requirements, power consumption, and thermal bottlenecks~\cite{amrouch2020npu}.

Approximate computing~\cite{han2013approximate} has recently emerged as the dominant paradigm that trades quality loss for energy and performance gains. A great amount of DNN operations can tolerate some degree of approximation~\cite{mrazek2019alwann,tasoulas2020weight}, and since the majority of DNN inference is spent on convolution and matrix multiplication operations, the design of approximate MAC units has attracted significant interest. 
Particularly, the majority of research has been focused on the design of approximate multipliers~\cite{axddnnsurvey2022}, 
as they are the most complex components of the MAC units and dominate energy consumption inside the unit. However, such multipliers are not application specific. They have been designed with static approximation and, once deployed, the generated circuits cannot adapt to input changes, causing serious integration problems. 

To balance this energy gain/accuracy trade-off dilemma, approximate accelerators have been proposed following two main architectural designs:
\begin{inparaenum}[(i)]
    \item tile-based MAC units, and
    \item MAC units with reconfigurable approximate multipliers.
\end{inparaenum}
Regarding the first method, instead of having a mesh of MAC units, the DNN accelerator is organized as a mesh
of tiles. Each tile hosts a combination of exact and multiple static approximate multipliers~\cite{mrazek2019alwann}. Then, based on the DNN and the system constraints, an analysis is performed to map the DNN layers on the appropriate multiplier (exact or approximate), while the rest are power-gated.
Since each layer of a DNN has different error sensitivity, the layer-to-static multiplier mapping is not a trivial problem; the design space is so big that exhaustive exploration is prohibited~\cite{mrazek2019alwann}. Additionally, such a coarse-grain \emph{layer-based} mapping of approximation limits the energy gains~\cite{tasoulas2020weight}. The second method is more fine-grain, flexible, and focuses on the design of 
reconfigurable approximate multipliers~\cite{tasoulas2020weight,spantidi2021positive}. Such multipliers sacrifice area to support multiple approximation modes with different introduced errors and are able to change the level of introduced error at run-time, based on the the weight values of each layer, to offer more fine-grain control. However, the main problem regarding their utilization lies in how to decide which approximation mode of the multiplier will be triggered for the different weight values in order to keep DNN accuracy within specific thresholds. 
Overall, the difficulty lies in the fact that there is no \emph{systematic approach to explore different weight-to-approximation mappings} as previous works are based on hybrid methods~\cite{zervakis2020design,mrazek2019alwann} and require constant manual tuning~\cite{tasoulas2020weight,spantidi2021positive}. 
Additionally, they 
target only the average accuracy of a dataset, which can be misleading in some cases as we show is Section~\ref{sec:motivation}. 
In Section~\ref{sec:evaluation} we will also show how the energy gains achieved from state-of-the-art related works at DNN accelerator level are fairly suboptimal due to inefficient exploitation of the underlying hardware.

To create more efficient mappings, while automating and enabling the systematic exploration of the introduced error, we can express the properties of the approximate accelerators using Signal Temporal Logic (STL)~\cite{HoxhaDF17sttt}, a specification
formalism to express system properties. Once an STL formula is expressed, a robustness analysis~\cite{bartocci2018specification} is used to evaluate in which cases the properties hold true. This way, we can check the system robustness under different STL expressions which allows systematic exploration, but is not scalable as each expression captures a small part of the exploration. Parametric Signal Temporal Logic (PSTL) \cite{asarin2011parametric} extends STL by replacing threshold constants with parameters that get to be inferred.

In this paper, we address the aforementioned problems regarding
weight-to-approximation mapping for energy-efficient DNN inference on approximate accelerators. Particularly, we present a unified framework that utilizes PSTL to express accuracy constraints and find weight-to-approximation mappings for the weights of a DNN on approximate multipliers, so as the constraints are satisfied and the energy consumption is minimized.
 The contributions of the proposed framework are:
\begin{itemize}[leftmargin=*]
	\item we propose a novel framework which, based on appropriate constructed formal expressions and input formulation, utilizes robustness metrics to achieve energy-efficient approximation mappings;
	\item we systematically explore the properties of approximate accelerators when multipliers with different introduced errors are employed, in correlation with their utilization under specific accuracy thresholds;
	\item we show that although the state of the art satisfies tight but coarse constraints, it fails to satisfy more fine-grain ones. To the best of our knowledge, this is the first work that allows fine-grain exploration of approximate accelerator error properties under multiple accuracy constraints, while avoiding manual tuning and retraining or weight tuning; and
	\item we evaluate the resulting mapping combinations against state-of-the-art mapping methodologies and compare our energy gain findings when considering both fine- and coarse-grain accuracy requirements. 
\end{itemize}

\vspace{-5pt}

\section{Related work}

Approximate computing has been heavily utilized on DNN inference~\cite{axddnnsurvey2022}. Many works present mapping methodologies that balance out the computation accuracy-power consumption trade-off, and
recent research has focused on the design and utilization of approximate multipliers on DNN inference. 

The work in~\cite{vasicek2019automated} deals with the automated design of application-specific approximate circuits using Cartesian Genetic Programming (CGP), employing a weighted mean error distance metric to guide the circuit approximation process.
In~\cite{ansari2019improving} the authors employ designs of CGP-based multipliers to achieve energy and area savings.  
The authors in~\cite{sarwar2018energy} utilize the notion of computation sharing and propose a compact and energy-efficient Multiplier-less Artificial Neuron. However, all works in ~\cite{vasicek2019automated,ansari2019improving,sarwar2018energy} require DNN retraining to retrieve some of the accuracy loss, which cannot always be applied~\cite{convar:dac2021}.

To avoid retraining, the work in~\cite{mrazek2019alwann} employs a layer-based (coarse) approximation method where each static multiplier is taken from~\cite{mrazek2017evoapprox8b} and it can be assigned to a distinct layer. The resulting multiplier selection problem is solved by a multi-objective optimization algorithm, however the applied mapping is layer-based which leaves room for improvement.
The work in~\cite{mrazek2020libraries} extends the library in~\cite{mrazek2017evoapprox8b} using CGP optimization to generate approximate multipliers, achieving significant energy gains with minimal accuracy loss for less complex networks.
In~\cite{hanif2019cann}, the authors present a compensation module that reduces energy consumption, but the additional accumulation row in the MAC array increases the computational latency.
Similarly, in~\cite{convar:dac2021} an additional MAC array column is used to predict and compensate the approximate multiplications error.
The work in~\cite{guo2020reconfigurable} proposes a reconfigurable approximate multiplier for quantized CNN applications that enables the reusability of resources for multiplications of different precisions.
In~\cite{zervakis2020design} the authors generate approximate multipliers with the capability of run-time reconfigurable accuracy.
The work in~\cite{tasoulas2020weight} uses~\cite{zervakis2020design} to generate low-variance approximate reconfigurable multipliers, and also proposes a fine-grain weight-oriented mapping methodology to achieve high energy gains with small accuracy losses.
Additionally,~\cite{spantidi2021positive} presents a dynamically configurable approximate multiplier that comprises an exact, a positive error, and a negative error mode, and proposes a filter-oriented approximation method that achieves high energy gains under different accuracy constraints, but the exploration time becomes unmanageable for larger DNNs and datasets.

A canonic sign digit-based approximation methodology for representing the filter weights of pre-trained Convolutional DNNs is presented in~\cite{riaz2020caxcnn}.
The work in~\cite{hammad9cnn} proposes an architecture comprising a pre-processing precision controller at the system level and approximate multipliers of varying precisions on the Processing Element (PE) level. The authors in~\cite{park2021design} utilize different approximate multipliers in an interleaved way to reduce the energy consumption of MAC-oriented signal processing algorithms.
However, both of the works~\cite{hammad9cnn, park2021design} target 16-bit inference, while modern DNN accelerators are mainly using 8-bit precision~\cite{jouppi2017datacenter}. Towards the optimization of DNN accelerators, the work in~\cite{zhang2022full} presents a full-stack accelerator search technique which improves the performance per thermal design power ratio. The work in~\cite{kosaian2021boosting} transforms convolutional and fully-connected DNN layers to achieve higher performance in terms of FLOPs/sec.

When compared to related works, our proposed approach differentiates in the following points:
\begin{inparaenum}[(1)]
	\item given a reconfigurable multiplier, we systematically produce approximation mappings through the exploration of the error properties of approximate accelerators,
	\item we employ multiple accuracy constraints for a given DNN and dataset,
	\item our proposed framework can receive any trained and quantized DNN as input and does not require retraining.
\end{inparaenum}

\vspace{-5pt}

\section{Motivation}\label{sec:motivation}

This section contains a motivational example that shows the necessity of automatic and fine-grain exploration of the error properties of approximate accelerators in order to generate efficient mappings. 
Overall, we focus our analysis on state-of-art approximate multipliers and mapping methodologies~\cite{tasoulas2020weight,mrazek2019alwann,spantidi2021positive,zervakis2020design}. 
Specifically, the works in~\cite{tasoulas2020weight,spantidi2021positive,zervakis2020design} propose reconfigurable multiplier designs that comprise three modes of operation, each one introducing varying levels of error and energy gains. These works also present mapping methodologies and specifically the works in~\cite{mrazek2019alwann,zervakis2020design} present layer-wise approximation mappings where each layer is entirely mapped to a different multiplier or multiplier mode respectively. To combat minimal energy gains achieved by layer-wise approaches, the works in~\cite{tasoulas2020weight,spantidi2021positive} propose fine-grain weight-oriented methodologies to decide which approximation modes of the respective reconfigurable multiplier will be used for each weight value in each layer of the DNN.

We argue that the existing mapping approaches are inadequate for the following reasons:
\begin{inparaenum}[(i)]
	\item they may result in resource under-utilization due to biased decisions;
	\item they target only average accuracy, which can be misleading; and
	\item they ignore big accuracy drops on specific dataset batches.
\end{inparaenum}

\ul{\textbf{Resource under-utilization due to biased decisions:}}
With the term \emph{utilization} we refer to the amount of times each distinct multiplier mode is being used in a mapping.
The methods in~\cite{tasoulas2020weight,spantidi2021positive} both perform weight-based mapping of approximation by employing the concepts of layer significance and weight magnitude. 
For instance, the authors in~\cite{tasoulas2020weight} presented LVRM (Low-Variance Reconfigurable Multiplier), an approximate reconfigurable multiplier that supports three operation modes:
\begin{inparaenum}[(i)]
	\item $LVRM0$, which triggers exact multiplications;
	\item $LVRM1$, which introduces low error and small energy gains contrary to $LVRM0$; and 
	\item $LVRM2$, which introduces greater error with larger energy gains than $LVRM1$.
\end{inparaenum}	
The method in\cite{tasoulas2020weight} tries to initially identify which layers are more resilient to error, and map their weights entirely to the most aggressive approximate mode $LVRM2$. Then, the weights of the remaining layers are mapped first to $LVRM2$, then to $LVRM1$, and finally to $LVRM0$ based on some experimentally derived ranges around the value zero, requiring manually tuning as each layer has different weight distribution.
Even though this method produces considerable energy savings, it is not scalable and results in \emph{resource under-utilization}.
By mapping complete layers to $LVRM2$, \cite{tasoulas2020weight} introduces significant error and makes the DNN susceptible to further approximation.
Thus, the utilization of the $LVRM1$ mode is reduced. As an example, their methodology on the ResNet20 and CIFAR-10 dataset, for a $0.5\%$ accuracy drop threshold, produces a mapping where the $22\%$ of the total multiplications is assigned to $LVRM0$, only $2\%$ to $LVRM1$, and the vast majority of $76\%$ to $LVRM2$. 
Thus, one of the approximate multiplier modes are barely utilized. 
Considering the sub-linear relation between induced error and energy reduction in approximate multipliers~\cite{mrazek2017evoapprox8b,ZervakisTVLSI2019}, we argue that approximating multiple layers with $LVRM1$ (i.e., moderate approximation) will deliver an energy reduction closer to linear and thus, potentially higher energy reduction than the one achieved by approximating only a few layers with $LVRM2$.

\begin{figure}
    \centering
    \resizebox{0.78\columnwidth}{!}{\includegraphics{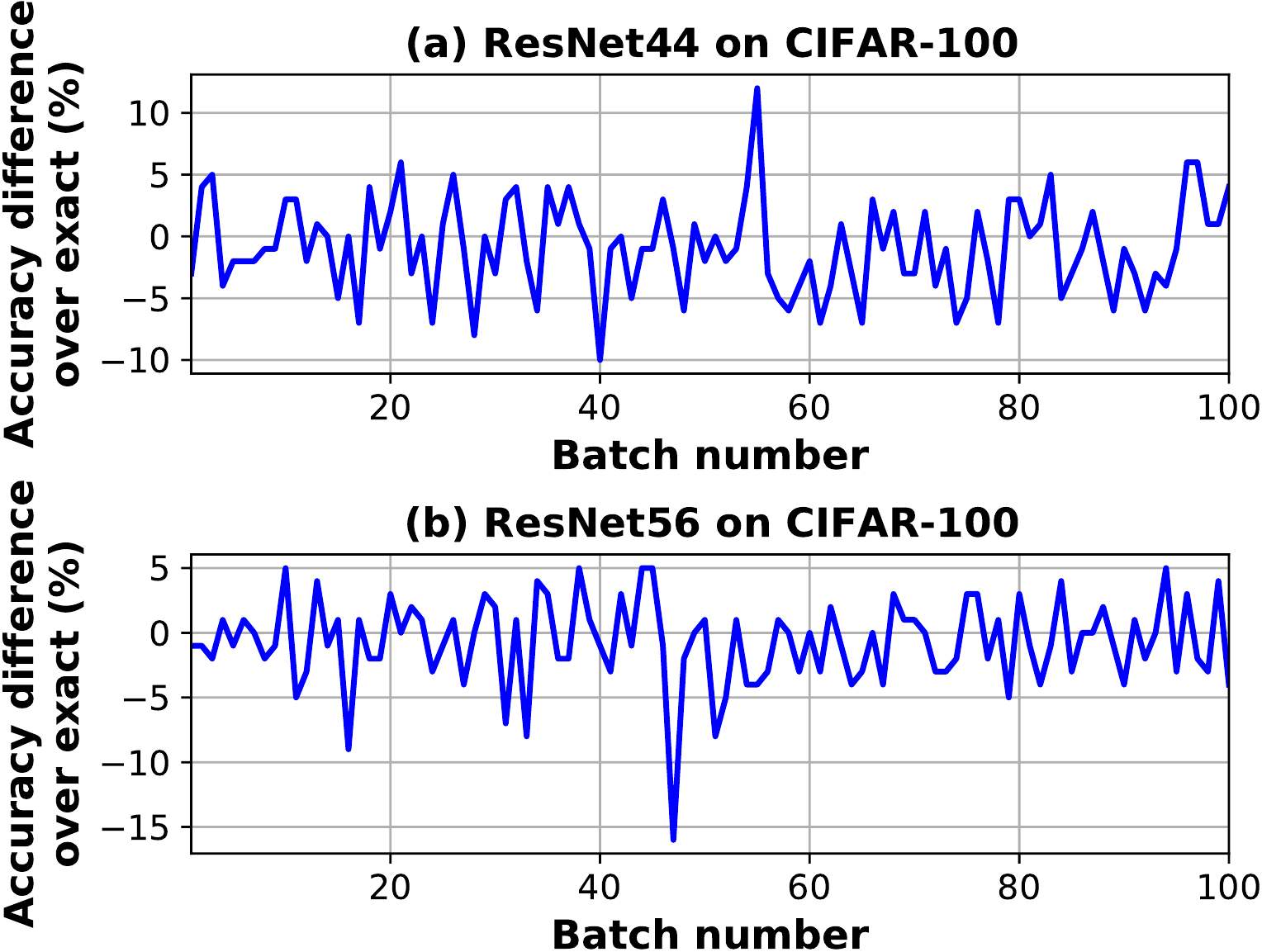}}
    \caption{(a) ResNet44 on CIFAR-100 with the multipliers in~\cite{mrazek2017evoapprox8b} and method in~\cite{mrazek2019alwann} and (b) ResNet56 on CIFAR-100 with the multiplier and method in~\cite{spantidi2021positive}.}
    \vspace{-15pt}
    \label{fig:motivation_eg}
\end{figure}

\ul{\textbf{Targeting only average accuracy can be misleading:}} Works that propose approximate multipliers and mapping methodologies~\cite{tasoulas2020weight,mrazek2019alwann, sarwar2018energy,spantidi2021positive,zervakis2020design}, evaluate their findings on the \emph{average accuracy} of a DNN over the target dataset. However, the introduction of error in computations does not affect all the batches of the dataset equally, creating, in many cases, great variations in the achieved accuracy. Such behavior is not acceptable when quality of service requirements need to be achieved over the entire dataset~\cite{dokhanchi2018evaluating}. Figure~\ref{fig:motivation_eg}(a) shows the inference accuracy differences of the ALWANN method~\cite{mrazek2019alwann} against the baseline (exact computations without error) for ResNet44 over the entire CIFAR-100 test dataset. Particularly, we split the 10,000 images into 100 equal batches and we show the accuracy differences for each one of them. Even though ALWANN had an overall accuracy drop of only 1\% over the entire dataset, when we take a deeper look into the achieved accuracy per batch, we can see cases where the accuracy drops as low as 10\% when compared to the exact operation (e.g., batch 40). Additionally, looking at the batches in which ALWANN achieved lower accuracy, over 20\% of them have an accuracy drop of more than 5\%, which is significant considering the strict constraint of 1\%. Similar behavior has been observed from the mappings produced by~\cite{tasoulas2020weight}.
Satisfying such fine-grain requirements could be posed in the form of \emph{queries}. For example, ``For the dataset batches that perform worse than the exact behavior, we want no more than 20\% of the cases to drop more than 5\% when we introduce approximation.'' In Section~\ref{sec:methodology} we show how to express formally such queries with PSTL and perform the corresponding fine-grain exploration. 

\ul{\textbf{Ignoring big accuracy drops on specific batches:}} Introducing approximation can also have an additional effect. Even though the variation of the achieved accuracy can be within specific values, there might be specific batches with very low accuracy. Figure~\ref{fig:motivation_eg}(b) shows the inference accuracy difference of the method in~\cite{spantidi2021positive} against the baseline (exact computations without error) for ResNet56 over the entire CIFAR-100 test dataset (10,000 images split into 100 equal batches).
Even though \cite{spantidi2021positive} achieves an overall accuracy drop of 1\% and satisfies the query presented in the previous observation, we can see that for batch 47 the accuracy drop is 16\%. Having such large drops during the inference phase could in some cases not be deemed acceptable, even if it is only for a small number of batches. 
For instance, such applications could be image recognition tasks in autonomous vehicles, where a steady stream of data is provided through different sensors (RADAR, LIDAR, etc)~\cite{dokhanchi2018evaluating, samal2020attention}. In such cases, it is important to study the accuracy for each block of this data stream instead of evaluating an NN over the final average accuracy.
Therefore, even more complex queries are needed to capture this behavior. For instance: ``For the dataset batches that perform worse than the exact behavior, we want no more than 20\% of the cases to drop more than 5\% and no case whatsoever to drop more than 15\% when we introduce approximation.'' Again, in Section~\ref{sec:methodology} we show how to express formally such queries with PSTL and perform the corresponding fine-grain exploration.

The analysis presented above shows the \emph{necessity to express complex queries in a formal way and explore solutions systematically without manual tuning. In that way, we will be able to produce flexible and scalable mappings, provide a more fine-grain control of the introduced error on the overall dataset, and support different levels of quality of service.}

\vspace{-5pt}

\section{Methodology}\label{sec:methodology}

In this paper, we present a methodology to systematically map DNN weights to approximation, under multiple and distinct objectives, using PSTL. By employing PSTL, we show that it is possible to describe more intricate properties of accelerators that comprise approximate multipliers. We additionally show that through PSTL we are allowed to formulate and solve optimization problems with respect to energy gains. 
The user provides an approximate multiplier, a DNN already trained and quantized to 8-bits and a dataset. The system we target in this work is a DNN accelerator comprising MAC units that utilize the given approximate multiplier. The output of the system is a single trajectory that captures the accuracy behavior of the given DNN for each batch of the given dataset. Specifically, the trajectory captures the \emph{accuracy drop} of the utilized approximate multiplier against the exact multiplier for each respective dataset batch. 
By considering this accuracy drop per batch for a given DNN as a trajectory, we can investigate more specific and fine-grain properties of the overall system and acquire more knowledge on the impact of approximate accelerators. 
Having defined the output trajectory of the accelerator, we can express a property query using PSTL. An example of such a query is ``For a given accelerator employing a reconfigurable multiplier and a given accuracy drop threshold, what is the maximum achievable energy gain we can achieve without violating the accuracy requirement?''. 
After expressing a PSTL query, the mapping exploration phase, also called \emph{parameter mining phase}, is triggered. Initially, the weights of the DNN are randomly assigned to approximation modes of the given reconfigurable multiplier.
The output accuracy trajectory is then analyzed for its robustness, which is then fed to a stochastic optimizer that decides on the next approximation mapping for each DNN layer. Essentially, the stochastic optimizer correlates the robustness value with per-layer approximation, and tries to find operating conditions that satisfy the defined PSTL query. The aim of the stochastic optimization step is to push the system's behavior to the constraint boundaries that are set through the PSTL queries. Once the exploration phase is completed, we build a Pareto-front of mined parameters where the PSTL query is guaranteed to be satisfied. 
\vspace{-5pt}

\subsection{Expressing system properties via Signal Temporal Logic}

As aforementioned, state-of-the-art mapping methodologies require manual tuning. For example, specific layers need to be selected based on their error resilience, and then further exploration is needed on individual weight value ranges~\cite{tasoulas2020weight,spantidi2021positive}. These methods mostly rely on experimental observations without \emph{systematic exploration}. How can queries like the ones described in Section~\ref{sec:motivation} be posed? How can we exploit such queries to infer system properties? In order to support such fine-grain exploration and automate the mapping procedure,
we utilize STL\cite{maler2004monitoring, HoxhaDF17sttt} and specifically
we build queries through PSTL~\cite{asarin2011parametric}, considering the accuracy of all the incoming batches over time as the output trajectory. From this point onward, we refer to such trajectories as \emph{signals}.

STL is a specification formalism used to express the properties of a given system in a compact way. To define the quantitative semantics of STL over arbitrary predicates, robustness is utilized as a quantitative measure on a given STL formula $\varphi$. Robustness indicates how far the signal is from satisfying or violating the defined STL specification~\cite{HoxhaDF17sttt}. 
An example of an STL expression is: ``Does the accuracy signal always remain above 98\% accuracy''. The syntax of STL comprises multiple Boolean operators 
but in this work, we only utilize the conjunction $\wedge$. Additional operators can be defined as syntactic abbreviations. In this work, we will be using the notion of the ``always'' operator~\cite{HoxhaDF17sttt} $\Box_\Ic \phi$, meaning that ``always during the interval ${\mathcal{I}}$,  $\phi$ should be true''. We consider ${\mathcal{I}} = [0,\infty)$, and therefore ${\mathcal{I}}$ can be dropped from the notation. 
These STL operators have been defined and used in literature to express temporal properties~\cite{maler2004monitoring, hoxha2018mining}.
Moreover, we extend ``always'' to a more relaxed operator  ${}^{X}\Box \phi$, where instead of demanding the entire specification $\phi$ to hold over the entire interval $\mathcal{I}$, we consider that $\phi$ is true if it holds just for $X\%$ of the signal values over the interval $\mathcal{I}$.

Using STL, it is possible to judge whether a signal satisfies or not a specified system property $\phi$. However, it is possible to also explore more elaborate system properties that a signal satisfies. Specifically, consider the aforementioned example of the STL query ``Does the accuracy signal always remain above 98\% accuracy?''. This query is essentially reduced to a ``yes or no'' problem: the signal either remains over 98\% over the entire interval $\mathcal{I}$ or not. Instead of pre-defining the 98\% accuracy value, we can leave it as a parameter $\theta$ to be mined. Therefore, the query can be rephrased as ``Which is the lowest accuracy value $\theta$ that the signal always satisfies?''. Such questions can be posed through the formal language PSTL~\cite{asarin2011parametric}. PSTL, an extension of STL, is a formal language used to express questions when we need to further explore the properties of a system instead of just determining whether an STL specification is satisfied or not. Parameter mining is the procedure of determining parameter values for PSTL formulas for which the specification is falsified. Therefore, parameter mining answers the question of which parameter ranges cause falsification~\cite{HoxhaDF17sttt}. 
Through the robustness metric, the parameter mining problem can be posed as an optimization problem~\cite{HoxhaDF17sttt}.

\vspace{-5pt}

\subsection{Parametric Signal Temporal Logic Queries}\label{sec:queries}

The employment of PSTL can assist in the exploration of DNN weight-to approximation mappings w.r.t. to maximizing a given parameter, which in our case is the \emph{energy savings} of the approximate accelerator. In this section, we show how we build incremental queries to capture energy gain values while keeping inference accuracy within multiple specific constraints: some of them more strict than others. Considering equal batches of our dataset as the stream of input data, we build the following initial query:

\begin{enumerate}
    \item[IQ1:] \textit{What is the maximum achieved energy gain $\theta$ such that, when applying approximation, any accuracy drop from the baseline is no more than $Accuracy_{thr}\%$ for X\% of the time?}
\begin{align*}
\varphi^{IQ1}[\theta] = & \square (Energy_{gain} \leq \theta) \implies \\
& {}^{X\%}\square (\text{$Accuracy_{diff}$} \leq Accuracy_{thr}\%)
\end{align*}
\end{enumerate}

In this initial query IQ1, values $X$ and $Accuracy_{thr}$ are defined by the user. For instance, in the motivation example presented in Figure~\ref{fig:motivation_eg}(a) and described in Section~\ref{sec:motivation} $X = 80\%$ and $Accuracy_{thr} = 5\%$. The baseline is the achieved accuracy without applying any approximation (i.e., exact computations). The parameter $\theta$ is at this stage unknown to the user and is left to be mined (Section~\ref{sec:methodology_mapping}). Note that, the $Accuracy_{diff}$ value refers to the \emph{accuracy difference per batch} $accuracy_{exact} - accuracy_{approximate}$ of each target DNN. In other words, we want to make sure that for all the input batches, in which approximation results in accuracy drop, this accuracy drop is no more than $Accuracy_{diff}$ for $X\%$ of these batches.

With the query IQ1, we are able to impose fine-grain constraints across all batches of the dataset regarding the variation of the accuracy drop. However, as we showed in Section~\ref{sec:motivation}, there are cases that even though the variation of the achieved accuracy is within specific values, there might be specific batches with very low accuracy. To that end we extend IQ1 as follows:
\begin{enumerate}
    \item[IQ2:] \textit{What is the maximum achieved energy gain $\theta$ such that, when applying approximation, any accuracy drop from the baseline is no more than $Accuracy_{thr}\%$ for X\% of the time, and no more than $Accuracy_{thr, total}$ at any time?}
\begin{align*}
\varphi^{IQ2}[\theta] = & \square (Energy_{gain} \leq \theta) \implies \\
& {}^{X\%}\square (\text{$Accuracy_{diff}$} \leq Accuracy_{thr}\%) ~\wedge \\
& \square (\text{$Accuracy_{diff}$} \leq Accuracy_{thr,total}\%)
\end{align*}
\end{enumerate}
Again in this case, the values of $X$, $Accuracy_{thr}$ and  $Accuracy_{thr,total}$ are defined by the user based on the needs of the application. 

Finally, since many related works on approximate reconfigurable multipliers take into consideration the average accuracy of each target DNN~\cite{mrazek2019alwann,tasoulas2020weight,spantidi2021positive}, we add it to query IQ2 to capture both fine-grain and coarse-grain accuracy information \emph{simultaneously}. Therefore:
\begin{enumerate}
    \item[IQ3:] \textit{What is the maximum achieved energy gain $\theta$ such that, when applying approximation, any accuracy drop from the baseline is no more than $Accuracy_{thr}\%$ for X\% of the time, no more than $Accuracy_{thr, total}$ at any time, and the average accuracy drop is below $Accuracy_{thr,avg}$?}
\begin{align*}
\varphi^{IQ3}[\theta] = & \square (Energy_{gain} \leq \theta) \implies \\
& {}^{X\%}\square (\text{$Accuracy_{diff}$} \leq Accuracy_{thr}\%) ~\wedge \\
& \square (\text{$Accuracy_{diff}$} \leq Accuracy_{thr,total}\%)~\wedge \\
& \square (\text{$Avg\_Accuracy\_Drop$} \leq Accuracy_{thr,avg}\%)
\end{align*}
\end{enumerate}
Once again, the values of $X$, $Accuracy_{thr}$, $Accuracy_{thr,total}$ and this time also $Accuracy_{thr,avg}$, are defined by the user. Concluding, in this section we used the initial query IQ1 to build the more elaborate query IQ3 in an attempt to profile more intricate accuracy properties for a given DNN. In our evaluation (Section~\ref{sec:evaluation}), we present the experimental results of different versions of the aforementioned queries, with different $X$, $Accuracy_{thr}$ and $Accuracy_{thr,avg}$ values.

\vspace{-5pt}

\subsection{Weight-to-approximation mapping}\label{sec:methodology_mapping}

In our methodology, we follow a layer-oriented approach to map specified multiplications to approximation. To that end, we utilize a stochastic optimizer under multiple accuracy constraints (e.g., query IQ 3), that iteratively
\begin{inparaenum}[(i)]
	\item takes as input the observed accuracy per batch and the estimated output energy of the system, and 
	\item produces as output a signal that describes the mapping of the DNN weights of each layer to the approximate modes of the multiplier. 
\end{inparaenum}

To formulate the problem and without loss of generality, we assume an accelerator whose MAC units are composed of reconfigurable approximate multipliers. Each reconfigurable approximate multiplier supports three operation modes:
\begin{inparaenum}[(i)]
	\item $M0$, which corresponds to the exact operation;
	\item $M1$, which introduces small error with small energy gains when compared to $M0$; and 
	\item $M2$,  which aggressively introduces greater error and achieves larger energy gains than $M1$.
\end{inparaenum}
Even though our method can be used for any number of approximation modes, we select three because previous research works have shown that the area overhead is not big and the control logic remains simple~\cite{tasoulas2020weight,spantidi2021positive}. 
Supposing that a DNN consists of $L$ layers ($l_1, l_2, \cdots, l_L$), then the outputs of the stochastic optimizer are two signals $V^{M2} =[v^{M2}_1, v^{M2}_2, \cdots, v^{M2}_L]$ and $V^{M1} =[v^{M1}_1, v^{M1}_2, \cdots, v^{M1}_L]$, where each element in both of them is a number between $0$ and $1$. Each element $v^{M2}_i$ represents the percentage of all the multiplications of layer $l_i$ that are chosen to be mapped on the approximate mode $M2$. Similarly, $v^{M1}_i$ is the percentage of all the multiplications of layer $l_i$ mapped on the approximate mode $M1$. Therefore, for any layer $l_i$, the percentage of the total layer's multiplications, that will be mapped to mode $M0$, will be $1-(v^{M1}_i + v^{M2}_i)$. Overall, the main goal of the stochastic optimizer is to observe the behavior of the DNN, in terms of accuracy and energy consumption, correlate it with the values of $V^{M2}$ and $V^{M1}$, and finally find appropriate values iteratively for the two vectors such that the constraints are satisfied.

\begin{figure}
    \centering
    \resizebox{0.75\columnwidth}{!}{\includegraphics{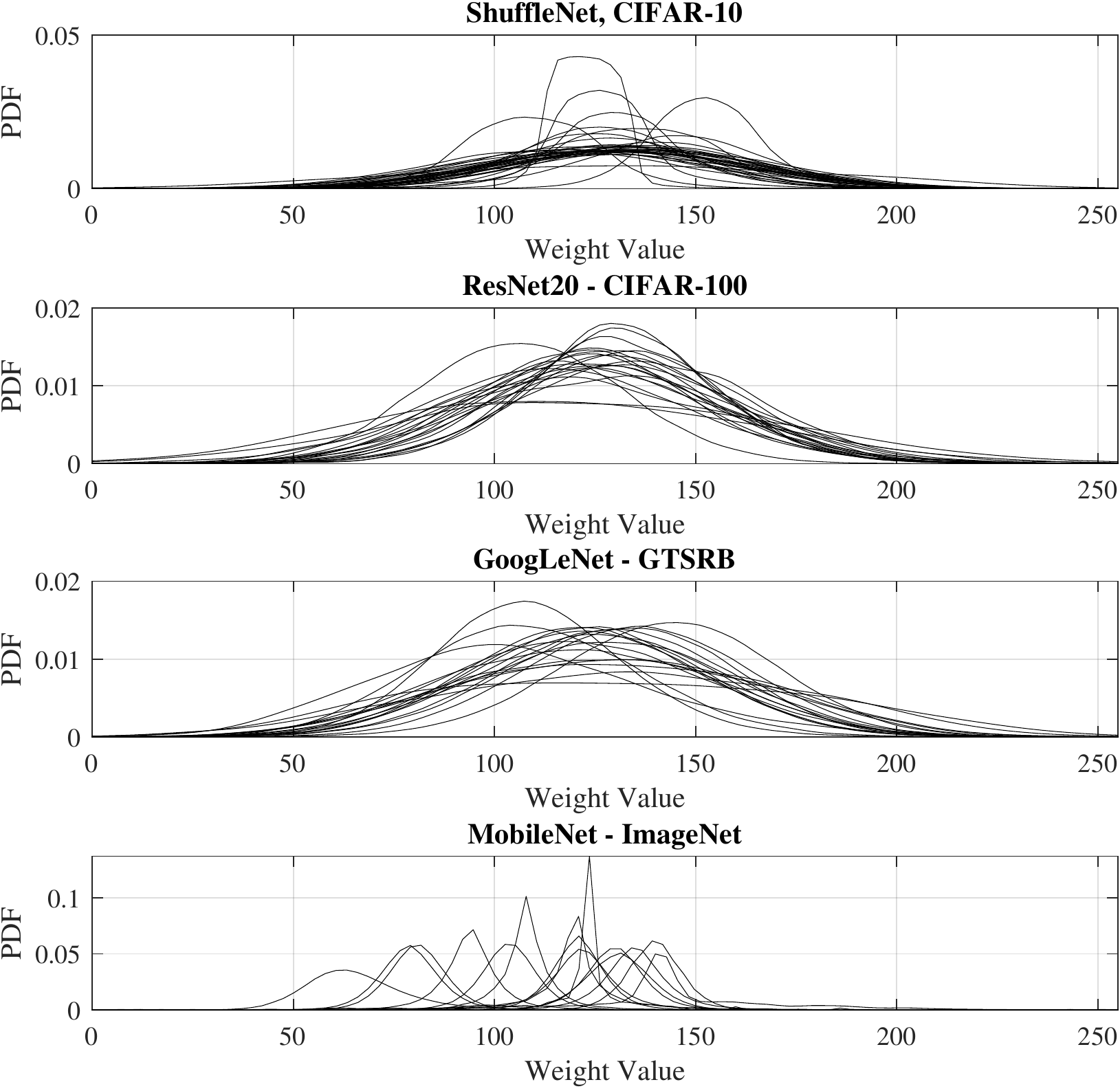}}
    \caption{The weight distribution for all layers on different networks (ShuffleNet, ResNet20, GoogLeNet, MobileNet) on different datasets (CIFAR-10, CIFAR-100, GTSRB, ImageNet). 8-bit quantization in [0,255] is used.}
    \vspace{-15pt}
    \label{fig:weight_distr}
\end{figure}

To guide the optimizer regarding which weight values would be most likely assigned to modes $M1$ and $M2$, we assign the different approximate modes to ranges around the median value of the weights for each DNN layer. We base this decision on the fact that in most cases the values of the weights of each layer are gathered around a centered value, featuring low dispersion as shown in~\cite{spantidi2021positive}. 
This way, the more aggressive $M1$ and $M2$ modes would be utilized more frequently, maximizing the potential for higher energy gains.
For completeness, Figure~\ref{fig:weight_distr} shows the weight distribution of all layers for four different networks (ShuffleNet, ResNet20, GoogLeNet, and MobileNet) on four different datasets (CIFAR-10, CIFAR-100, GTSRB, and ImageNet). 
From our analysis, it is observed that the vast majority of layers follow this principle having either a very distinguish peak or a more flattened behavior. However, there were no layers with multiple peaks regarding the distribution. 
To accommodate fine-grain mapping we follow the approach described in~\cite{tasoulas2020weight} where a 2-bit signal is used to select the multiplier mode, and a control unit that comprises 4 8-bit comparators, two AND gates and an OR gate to activate multiplier modes based on weight values (i.e. selected ranges). Overall, the hardware needed to support such fine-grain weight mapping to different multiplier modes using ranges results in a minimal area overhead of less than 3\%~\cite{tasoulas2020weight}.
Also, the number of control units is equal to the number of the MAC array rows and thus, it increases only linearly as the MAC array size increases (quadratically).

\begin{figure}
    \centering
    \resizebox{0.75\columnwidth}{!}{\includegraphics{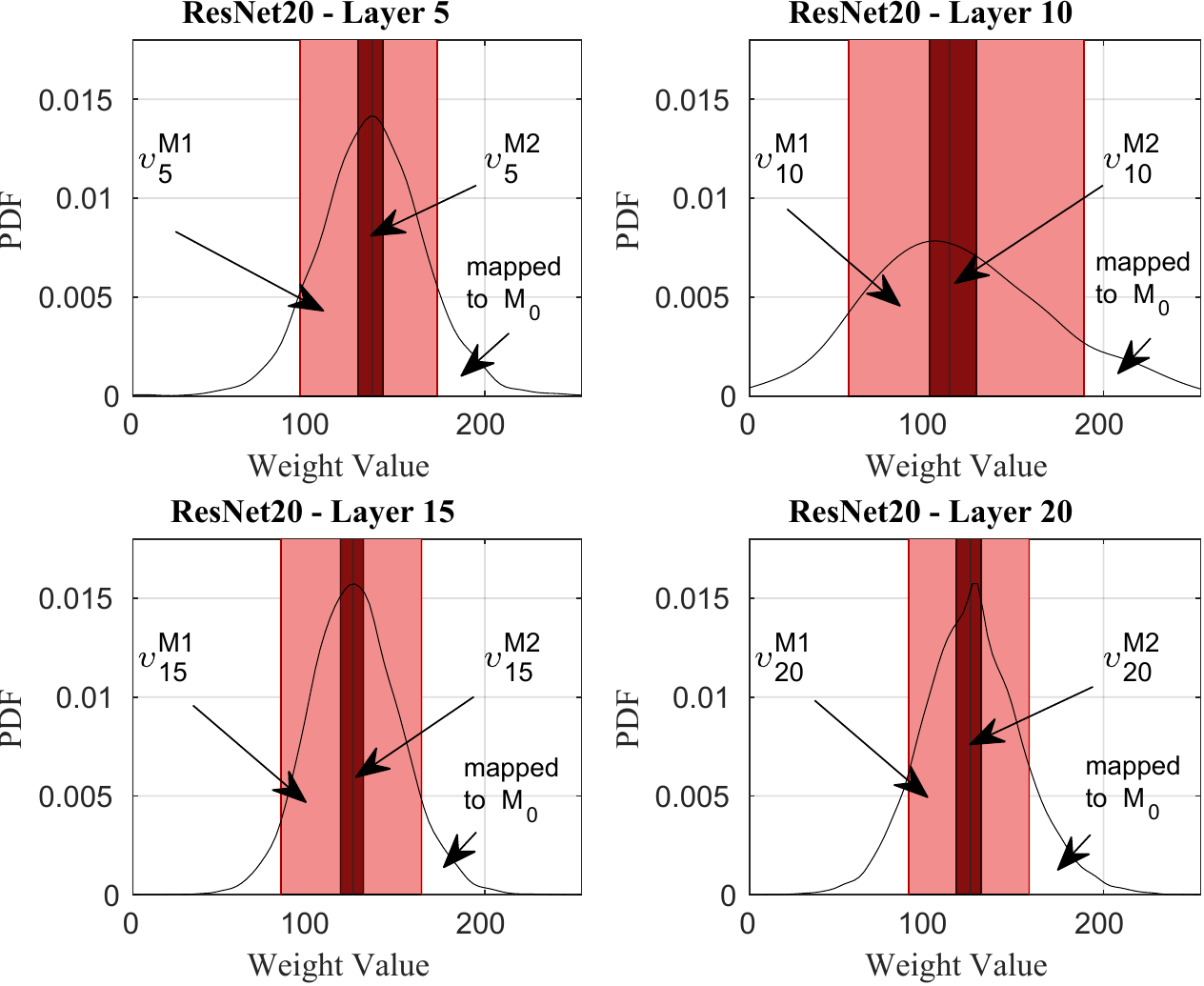}}
\caption{Mapping example for ResNet20 on CIFAR-10. Darker area indicates the mapping range of $M2$ mode, lighter area refers to $M1$ mapping and uncolored area refers to $M0$ mapping. 8-bit quantization in [0,255] is used.}
    \vspace{-10pt}
    \label{fig:mapping_eg}
\end{figure}

Regarding the values assigned to $V^{M2}$ and $V^{M1}$, Figure~\ref{fig:mapping_eg} shows an example of the weight distribution for four different layers of ResNet20 and how the proposed mapping is applied. The dark red area represents the weight values that are mapped to $M2$ ($v^{M2}_i$), while the lighter red area represents the weight values that are mapped to $M1$ ($v^{M1}_i$). The rest of the values that are not included in either colored patches are mapped to $M0$. By modifying the values of $V^{M2}$ and $V^{M1}$ the stochastic optimizer widens or narrows down the colored areas to find the best solution that satisfies the required constraints. 

\begin{figure}
    \centering
    \resizebox{0.75\columnwidth}{!}{\includegraphics{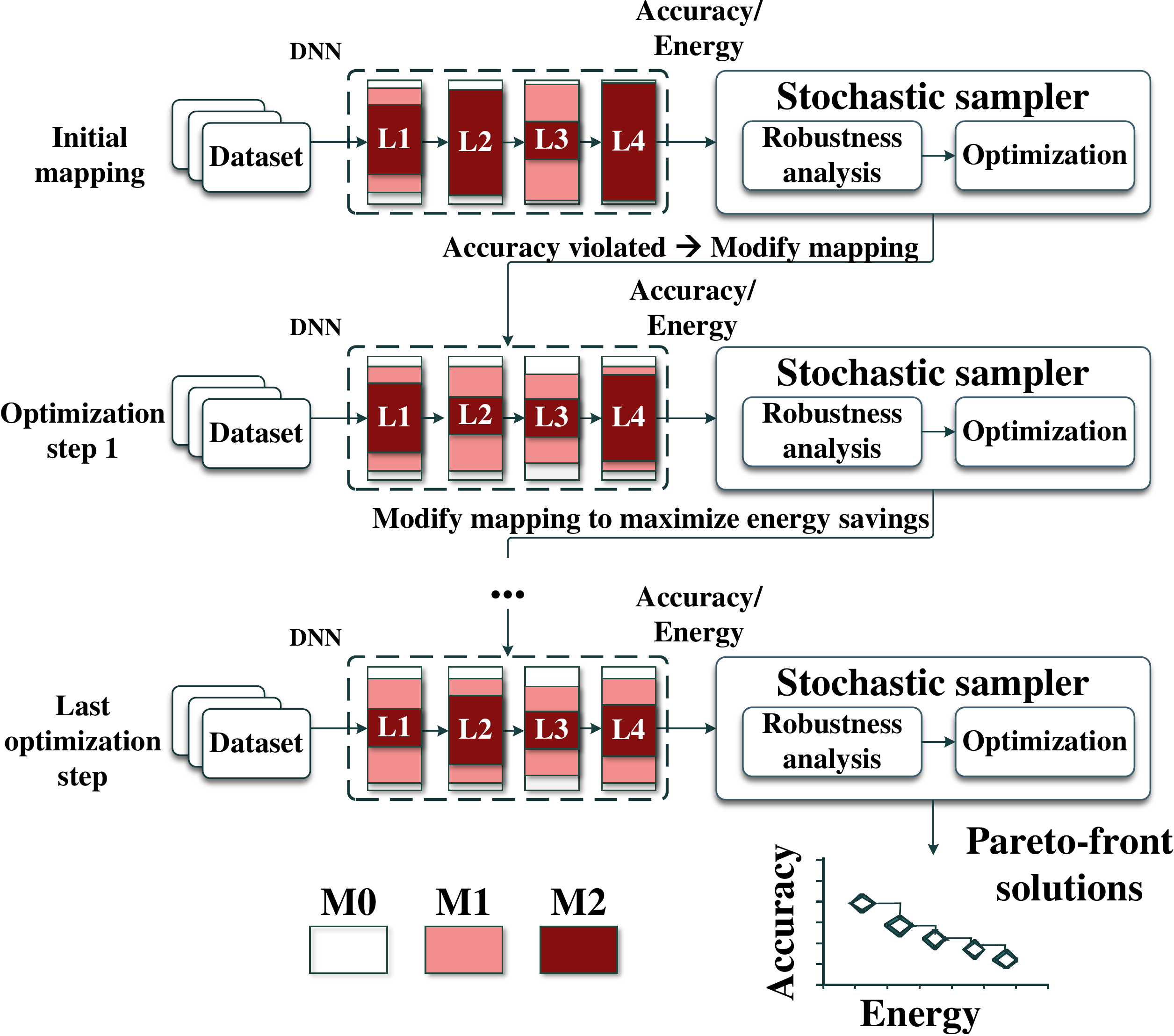}}
    \caption{Optimization steps for mapping DNN weights to approximate modes.}
    \label{fig:example}
    \vspace{-15pt}
\end{figure}

When the exploration is triggered on a query $\varphi^{IQ}[\theta]$, initially the $V^{M1}$ and $V^{M2}$ signals contain random values (Figure~\ref{fig:example}).
The output signal of the accelerator for the executed DNN, in terms of accuracy per batch, is then analyzed for its robustness, the stochastic optimizer correlates the robustness value with per layer approximation, and it alters the impact of the approximation through modifications on $V^{M1}$ and $V^{M2}$. In each optimization iteration, the stochastic optimizer aims to gradually minimize the analyzed robustness of the system based on its output. Therefore, the goal of the optimizer is to eventually tweak the $V^{M1}$ and $V^{M2}$ signals in a way that would satisfy all the given constraints described in the query (e.g., IQ1-IQ3). Regarding the utilized stochastic optimizer, we employ the Expected Robustness Guided Monte Carlo (ERGMC) algorithm, which is based on simulated annealing and is presented in~\cite{abbas2014robustness}. 
We set the number of control points to be equal to the amount of convolution layers of each target DNN, and evenly distribute them.
Overall, the stochastic optimizer aims to push the system's behavior as close as possible to the specified constraint boundaries. The parameter mining phase is completed after a predefined number of tests.
We then generate a Pareto-front in the parameter space, resulting from all the conducted tests. What we consider to be the final output of this phase is the mapping that corresponds to the maximum found value of the parameter $\theta$.

\begin{figure}
    \centering
    \resizebox{1\columnwidth}{!}{\includegraphics{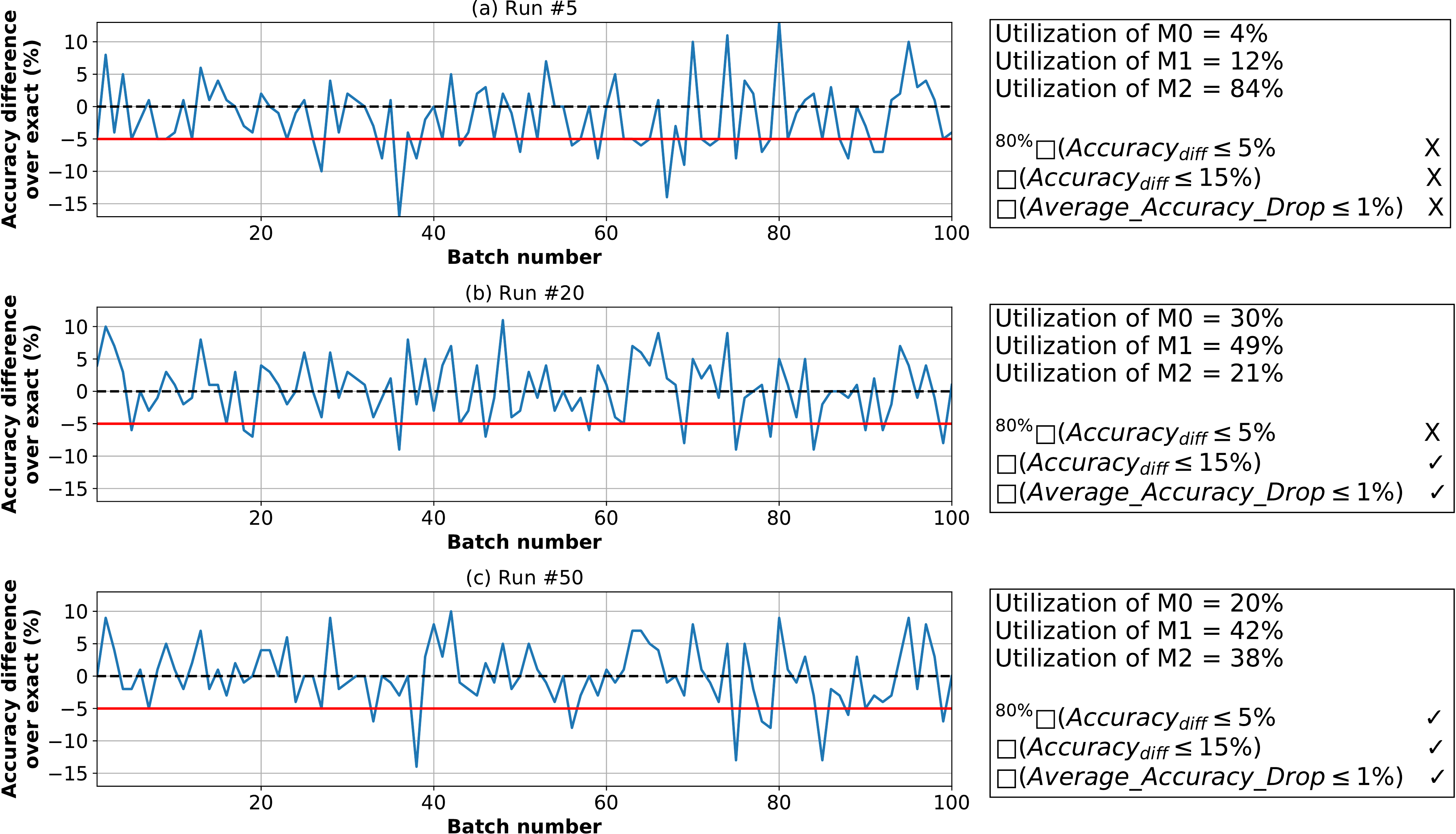}}
    \caption{A parameter mining example on GoogLeNet and CIFAR-100, showcasing different approximation mappings selected by the stochastic optimizer.}
    \vspace{-15pt}
    \label{fig:optimizer}
\end{figure}

Figure~\ref{fig:optimizer} shows an example of the parameter mining process on the GoogLeNet network on the CIFAR-100 dataset. We utilized LVRM~\cite{tasoulas2020weight} as the approximate multiplier which comprises three modes of operation namely $M0$, $M1$, and $M2$.
This example is based on the query IQ3 ($\varphi^{IQ3}[\theta]$) presented in Section~\ref{sec:queries}, where $X = 80\%$, $Accuracy_{thr} = 5\%$, $Accuracy_{thr,total} = 15\%$, and $Accuracy_{thr,avg} = 1\%$. Additionally, we set the maximum number of tests for the stochastic optimizer to $50$.
For each run, Figure~\ref{fig:optimizer} shows the achieved accuracy for each batch of the dataset (output signal), the corresponding utilization of the approximate modes across all the layers of the DNN, and the constraints of the query that were satisfied.
In the very first run of the parameter mining phase all weights are assigned to an approximate mode randomly. By the fifth run of the parameter mining phase (Figure~\ref{fig:optimizer}(a)), $84\%$ of the weights are mapped on $M2$, $12\%$ to $M1$ and the rest to $M0$, since the stochastic optimizer correlates energy gains to $V^{M1}$ and $V^{M2}$ values, pushing for more energy savings through increased $M2$ utilization.
The robustness for this output signal is negative as it falsifies all constraints. Thus, the optimizer modifies the values of $V^{M1}$ and $V^{M2}$ to correlate the robustness value with the per layer introduced approximation. By letting the optimizer run for a few more tests, Figure~\ref{fig:optimizer}(b) shows the output signal at run \#20 and the corresponding information about the utilization of the approximate modes and the satisfaction of the constraints. Since the initial high utilization of $M2$ resulted in very low robustness, the optimizer tried to utilize the $M1$ mode more to make the robustness value higher, as it introduces less error. At this run, the utilization of $M1$ is 49\%, of $M2$ 21\% and of $M0$ 30\%. This mapping satisfies two of the constraints but still not all of them, therefore the robustness of the system is still negative. By letting the optimizer continue, in the last run \#50 (Figure~\ref{fig:optimizer}(c)), we see that all constraints are satisfied and the optimizer managed to find a balance for the utilization of the approximate modes and thus, the robustness of the system is now positive. 
Overall, the robustness value is an indication of how ``far'' or ``near'' the signal is from satisfying the initially set accuracy requirements. The robustness is evaluated on the accuracy signal based on the PSTL requirements, and is then utilized by the stochastic optimizer to select the next approximation mappings, which lead to $\theta$ parameter ranges.
\vspace{-5pt}

\section{Evaluation}\label{sec:evaluation}

We aim to showcase the strengths of the proposed method in terms of
\begin{inparaenum}[1)]
	\item efficient mappings with higher energy savings than previous methods;
	\item efficient utilization of all approximate modes;
	\item automatic and scalable fine-grain exploration.
\end{inparaenum}
Regarding the energy consumption, the MAC units are described in Verilog RTL, synthesized using Synopsys Design Compiler and mapped to a 7nm technology library. Mentor Questasim is used for post-synthesis timing simulations and the switching activity of the MAC units is captured through 1 million randomly generated inputs, which is then fed to Synopsys PrimeTime to calculate power consumption. Additionally, we bridge \staliro~\cite{HoxhaDF17sttt}, a toolbox for temporal logic falsification, with the Tensorflow machine learning library, in which we overrode the convolution layers and replaced the exact multiplications with the respective approximate ones~\cite{mrazek2019alwann}. Finally, we use the stochastic optimizer in \staliro to solve the PSTL query and acquire the approximation mappings.

We consider seven different PSTL queries and evaluate our findings against the mapping methods presented in LVRM~\cite{tasoulas2020weight} and ALWANN~\cite{mrazek2019alwann} across seven DNNs quantized to 8 bits: GoogleNet~\cite{szegedy2015going}, MobileNetv2~\cite{sandler2018mobilenetv2}, ResNet20~\cite{he2016deep}, ResNet32~\cite{he2016deep}, ResNet44~\cite{he2016deep}, ResNet56~\cite{he2016deep}, and ShuffleNet~\cite{zhang2018shufflenet}. All the aforementioned DNNs were trained on three datasets: CIFAR-10, CIFAR-100, and GTSRB each of which considers $32\times32$ input image size. We additionally evaluate our method on the Imagenet dataset ($224 \times 224$ image size) on the following four DNNs: InceptionV3~\cite{szegedy2016rethinking} (rescales images to $299 \times 299$), MobileNet~\cite{howard2017mobilenets}, NasNet~\cite{zoph2018learning}, and VGG16~\cite{simonyan2014very}.
We also utilized the method presented in~\cite{spantidi2021positive}, however due to the in-depth per filter search, the exploration time for the Imagenet dataset was extremely high. We use 25\% of each dataset during the optimization phase. We provide more information about execution time in Section~\ref{sec:eval_exec_time}.

\vspace{-5pt}
\subsection{Considered PSTL queries}

\begin{table*}[]
\caption{All considered queries expressed in PSTL with their respective description, where $Accuracy_{thr,avg} = \{0.5\%, 1\%, 2\%\}$.}
\scriptsize
\label{tab:queries}
\centering
\resizebox{0.85\textwidth}{!} 
{ 
\begin{tabular}{ll} \toprule
\multicolumn{2}{c}{\textbf{What is the maximum achieved energy gain $\theta$ during inference such that:}} \\ \toprule \toprule
$\varphi^{Q1}[\theta] = \square (Energy_{gain} \leq \theta) \implies  {}^{40\%}\square (\text{$Accuracy_{diff}$} \leq 3\%) \wedge$           & any per batch accuracy drop is less than 3\% for 40\% of the batches, and          \\
$\hspace{11.9mm} \square (\text{$Accuracy_{diff}$} \leq 15\%) \wedge$           & the per batch accuracy drop is less than 15\% at any time, and                     \\
$\hspace{11.9mm} \square (\text{Avg\_Accuracy\_Drop} \leq Accuracy_{thr,avg})$          & the average accuracy drop is less than $Accuracy_{thr,avg}$.                            \\ \midrule
$\varphi^{Q2}[\theta] = \square (Energy_{gain} \leq \theta) \implies  {}^{60\%}\square (\text{$Accuracy_{diff}$} \leq 3\%) \wedge$           & any per batch accuracy drop is less than 3\% for 60\% of the batches, and          \\
$\hspace{11.9mm}\square (\text{$Accuracy_{diff}$} \leq 15\%) \wedge$           & the per batch accuracy drop is less than 15\% at any time, and                     \\
$\hspace{11.9mm}\square (\text{$Avg\_Accuracy\_Drop$} \leq Accuracy_{thr,avg})$           & the average accuracy drop is less than $Accuracy_{thr,avg}$.                            \\ \midrule
$\varphi^{Q3}[\theta] = \square (Energy_{gain} \leq \theta) \implies  {}^{80\%}\square (\text{$Accuracy_{diff}$} \leq 3\%) \wedge$           & any per batch accuracy drop is less than 3\% for 80\% of the batches, and          \\
$\hspace{11.9mm}\square (\text{$Accuracy_{diff}$} \leq 15\%) \wedge$           & the per batch accuracy drop is less than 15\% at any time, and                     \\
$\hspace{11.9mm}\square (Avg\_Accuracy\_Drop \leq Accuracy_{thr,avg})$           & the average accuracy drop is less than $Accuracy_{thr,avg}$.                            \\ \midrule
$\varphi^{Q4}[\theta] = \square (Energy_{gain} \leq \theta) \implies  {}^{40\%}\square (\text{$Accuracy_{diff}$} \leq 5\%) \wedge$           & any per batch accuracy drop is less than 5\% for 40\% of the batches, and          \\
$\hspace{11.9mm}\square (\text{$Accuracy_{diff}$} \leq 15\%) \wedge$            & the per batch accuracy drop is less than 15\% at any time, and                     \\
$\hspace{11.9mm}\square (Avg\_Accuracy\_Drop \leq Accuracy_{thr,avg})$           & the average accuracy drop is less than $Accuracy_{thr,avg}$.                            \\ \midrule
$\varphi^{Q5}[\theta] = \square (Energy_{gain} \leq \theta) \implies  {}^{60\%}\square (\text{$Accuracy_{diff}$} \leq 5\%) \wedge$           & any per batch accuracy drop is less than 5\% for 60\% of the batches, and          \\
$\hspace{11.9mm}\square (\text{$Accuracy_{diff}$} \leq 15\%) \wedge$            & the per batch accuracy drop is less than 15\% at any time, and                     \\
$\hspace{11.9mm}\square (Avg\_Accuracy\_Drop \leq Accuracy_{thr,avg})$           & the average accuracy drop is less than $Accuracy_{thr,avg}$.                            \\ \midrule
$\varphi^{Q6}[\theta] = \square (Energy_{gain} \leq \theta) \implies  {}^{80\%}\square (\text{$Accuracy_{diff}$} \leq 5\%) \wedge$           & any per batch accuracy drop is less than 5\% for 80\% of the batches, and          \\
$\hspace{11.9mm}\square (\text{$Accuracy_{diff}$} \leq 15\%) \wedge$            & the per batch accuracy drop is less than 15\% at any time, and                     \\
$\hspace{11.9mm}\square (Avg\_Accuracy\_Drop \leq Accuracy_{thr,avg})$           & the average accuracy drop is less than $Accuracy_{thr,avg}$.                            \\ \midrule \midrule
$\varphi^{Q7}[\theta] = \square (Energy_{gain} \leq \theta) \implies$           &                                                                                    \\
$\hspace{11.9mm}\square (Avg\_Accuracy\_Drop \leq Accuracy_{thr,avg})$           & the average accuracy drop is less than $Accuracy_{thr,avg}$.  \\ \bottomrule           
\end{tabular}
}
\vspace{-15pt}
\end{table*}

Our initial motivation was to express intricate properties of systems through PSTL and find energy-efficient approximation mappings given any approximate reconfigurable multiplier. In Section~\ref{sec:queries}, we presented the queries we considered to describe said properties. In our evaluation, we use variations of the initially presented queries to build the Queries Q1-Q7 as shown in Table~\ref{tab:queries}. We constructed queries that aim to capture different levels of requirements, with some queries being more relaxed than others. Note that, as mentioned in Section~\ref{sec:queries}, there are some user-defined variables: $X$, $Accuracy_{thr}$, $Accuracy_{thr,total}$ and $Accuracy_{thr,avg}$. For all of the considered queries in this evaluation, $Accuracy_{thr,total}$ is set to be 15\%. This $Accuracy_{thr,total}$ value was selected based on the assumption that such a big accuracy drop should be the maximum drop per batch that can be tolerated.
The queries shown in Table~\ref{tab:queries} are split into three main parts. 

\ul{\textbf{Strict fine-grain constraints (Q1-Q3):}} For the first three queries Q1-Q3, we set the maximum acceptable accuracy drop per batch from the baseline to be $Accuracy_{thr} = 3\%$ and we set this requirement to hold for $X = \{40\%, 60\%, 80\%\}$ of these batches. Requiring this specification to hold for $40\%$ of the batches (Q1) is a less aggressive approach compared to requiring it to hold for $80\%$ of the batches (Q3). Overall these three queries are the most strict considered in this evaluation.

\ul{\textbf{Relaxed fine-grain constraints (Q4-Q6):}} The next three queries Q4-Q6 in Table~\ref{tab:queries} are a more relaxed version of the previously analyzed Q1-Q3 queries. We variate $X\%$ in the same way, i.e., $X = \{40\%, 60\%, 80\%\}$ but this time the per batch acceptable accuracy drop threshold is larger and set to $Accuracy_{thr} = 5\%$. With queries Q4-Q6, we wanted to allow more room for aggressive approximation, by slightly increasing the acceptable accuracy drop per batch. Similar to the previous triad of queries, there is a gradient strictness as we move from query Q4 to Q6 caused by the $X\%$ value.

\ul{\textbf{No fine-grain constraints (Q7):}} Q7 is the last query considered in this evaluation and is the most relaxed among all. We did not impose any constraints per batch and we only set the coarse-grain requirement of $Accuracy_{thr,avg}$ to hold. We included this query in our evaluation since it is the same requirement enforced by previous works, considering only average accuracy drop~\cite{tasoulas2020weight,mrazek2019alwann,sarwar2018energy,spantidi2021positive,zervakis2020design}.

For all seven queries, we consider three different cases of $Accuracy_{thr,avg}$: $0.5\%, 1\%$ and $2\%$. The different combinations of $Accuracy_{thr}$, $Accuracy_{thr,total}$ and $Accuracy_{thr,avg}$ values provide a diverse coverage in terms of requirement strictness on DNN accuracy.
Note that, the proposed methodology can be scaled to any number of iterations and batch size. The following results are based on a number of 100 iterations and a batch size of 100 to stay consistent with the examples shown in Sections~\ref{sec:motivation} and \ref{sec:methodology_mapping}. We evaluated this work for smaller batch sizes with similar behavior. However, for smaller batch sizes the $Accuracy_{thr}$ and $Accuracy_{thr,total}$ values would require appropriate alterations.

\vspace{-5pt}

\subsection{Comparison with a weight-oriented mapping methodology}

\begin{figure}
\centering
  \resizebox{\columnwidth}{!}{\includegraphics{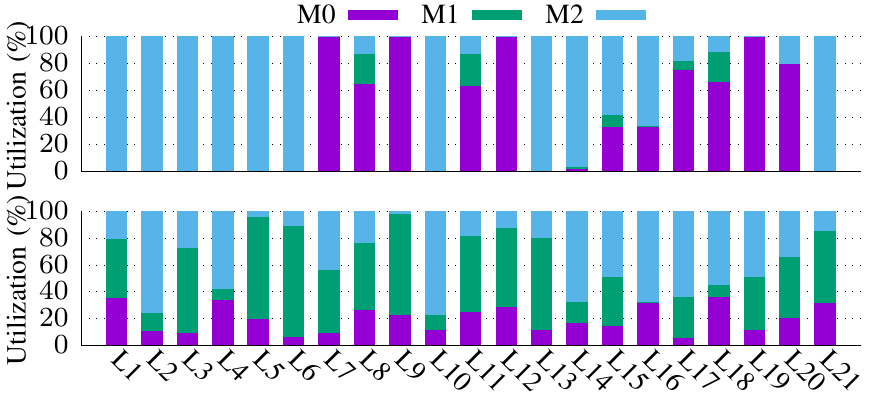}}
  \vspace{-10pt}
  \caption{Utilization of approximate modes, per layer, used by LVRM~\cite{tasoulas2020weight}(top) compared to our mapping (bottom) for ResNet20, CIFAR-10 dataset, Q7, and $1\%$ accuracy drop threshold.}
  \label{fig:lvrm_distribution}
  \vspace{-15pt}
\end{figure}

\begin{table}[h!]
\centering
\setlength{\tabcolsep}{3pt}
\caption{The queries LVRM~\cite{tasoulas2020weight} satisfies for all datasets and DNNs}
\scriptsize
\label{tab:lvrm_queries}
\resizebox{0.75\columnwidth}{!} 
{ 
\begin{tabular}{lccccccc}
\toprule
\textbf{CIFAR-10}    		& Q1      & Q2  & Q3  & Q4      & Q5  & Q6  & Q7            \\ \toprule \toprule
GoogLeNet            & 2\%     & \ding{55}   & \ding{55}   & 1\%, 2\% & \ding{55}   & \ding{55}   & \ding{52}       \\
MobileNetv2          & \ding{55}       & \ding{55}   & \ding{55}   & 1\%, 2\% & \ding{55}   & \ding{55}   & \ding{52} \\
ResNet20             & \ding{55}     & \ding{55}   & \ding{55}   & 2\%       & \ding{55}   & \ding{55}   & \ding{52} \\
ResNet32             & 2\% & \ding{55}   & \ding{55}   & 1\%, 2\%     & 2\% & 2\% & \ding{52} \\
ResNet44             & 2\%     & \ding{55}   & \ding{55}   & 2\%     & \ding{55}   & \ding{55}   & \ding{52} \\
ResNet56             & 1\%, 2\% & \ding{55}   & \ding{55}   & 1\%, 2\% & \ding{55}   & \ding{55}   & \ding{52} \\
ShuffleNet           & 2\%     & 2\% & 2\% & 1\%, 2\% & 2\% & 2\% & \ding{52} \\ \bottomrule \\
\toprule
\textbf{CIFAR-100}   & Q1  & Q2 & Q3 & Q4  & Q5 & Q6 & Q7      \\ \toprule \toprule
GoogLeNet   & \ding{55}   & \ding{55}  & \ding{55}  & 2\%   & \ding{55}  & \ding{55}  & \ding{52}       \\
MobileNetv2 & \ding{55}   & \ding{55}  & \ding{55}  & 2\%   & \ding{55}  & \ding{55}  & \ding{52}     \\
ResNet20    & 2\%   & \ding{55}  & \ding{55}  & 1\%, 2\% & \ding{55}  & \ding{55}  & \ding{52}     \\
ResNet32    & 2\%   & \ding{55}  & \ding{55}  & 1\%, 2\% & 2  & \ding{55}  & \ding{52}     \\
ResNet44    & \ding{55}   & \ding{55}  & \ding{55}  & 2\%   & \ding{55}  & \ding{55}  & \ding{52} \\
ResNet56    & 2\% & \ding{55}  & \ding{55}  & 2\%   & \ding{55}  & \ding{55}  & \ding{52} \\
ShuffleNet  & 2\%   & 1\%, 2\%  & \ding{55}  & 2\%   & 1\%, 2\%  & \ding{55}  & \ding{52} \\ \bottomrule \\
\toprule
\textbf{GTSRB}       & Q1  & Q2 & Q3 & Q4      & Q5 & Q6 & Q7      \\ \toprule \toprule
GoogLeNet   & \ding{55}   & \ding{55}  & \ding{55}  & \ding{52} & \ding{55}  & \ding{55}  & \ding{52} \\
MobileNetv2 & \ding{55}   & \ding{55}  & \ding{55}  & 1\%, 2\%     & 2\%  & \ding{55}  & \ding{52} \\
ResNet20    & 1\%, 2\% & \ding{55}  & \ding{55}  & 1\%, 2\%     & \ding{55}  & \ding{55}  & \ding{52}     \\
ResNet32    & 1\%, 2\% & \ding{55}  & \ding{55}  & \ding{52} & \ding{55}  & \ding{55}  & \ding{52} \\
ResNet44    & 2\%   & \ding{55}  & \ding{55}  & 1\%, 2\%     & \ding{55}  & \ding{55}  & \ding{52}     \\
ResNet56    & 1\%, 2\% & \ding{55}  & \ding{55}  & 1\%, 2\%     & \ding{55}  & \ding{55}  & \ding{52} \\
ShuffleNet  & \ding{55}   & \ding{55}  & \ding{55}  & 1\%, 2\%     & \ding{55}  & \ding{55}  & \ding{52} \\ \bottomrule \\
\toprule
\textbf{ImageNet}    & Q1  & Q2 & Q3 & Q4  & Q5 & Q6 & Q7      \\ \toprule \toprule
Inceptionv3 & \ding{55}   & \ding{55}  & \ding{55}  & 2\%   & \ding{55}  & \ding{55}  & \ding{52}     \\
Mobilenet   & 1\%, 2\% & \ding{55}  & \ding{55}  & 1\%, 2\% & \ding{55}  & \ding{55}  & \ding{52} \\
NasNet      & \ding{55}   & \ding{55}  & \ding{55}  & \ding{55}   & \ding{55}  & \ding{55}  & \ding{52}       \\
VGG16       & \ding{55}   & \ding{55}  & \ding{55}  & 2\%   & 2\%  & \ding{55}  & \ding{52}      \\ \bottomrule
\end{tabular}
}
\vspace{-15pt}
\end{table}

In this section, we present the benefits of our mapping approach against the weight-oriented mapping methodology presented in~\cite{tasoulas2020weight}. As aforementioned, \cite{tasoulas2020weight} presents LVRM, an approximate reconfigurable multiplier that supports three operation modes namely LVRM0 ($M0$), LVRM1 ($M1$), and LVRM2 ($M2$). $M0$ is the exact operation, $M1$ is the least aggressive approximate mode and $M2$ is the most aggressive one. LVRM additionally presents a four-step methodology that maps DNN weights to different multiplier modes based on the layer sensitivity. For the evaluation, we:

\begin{figure*}[h]
\centering
\includegraphics{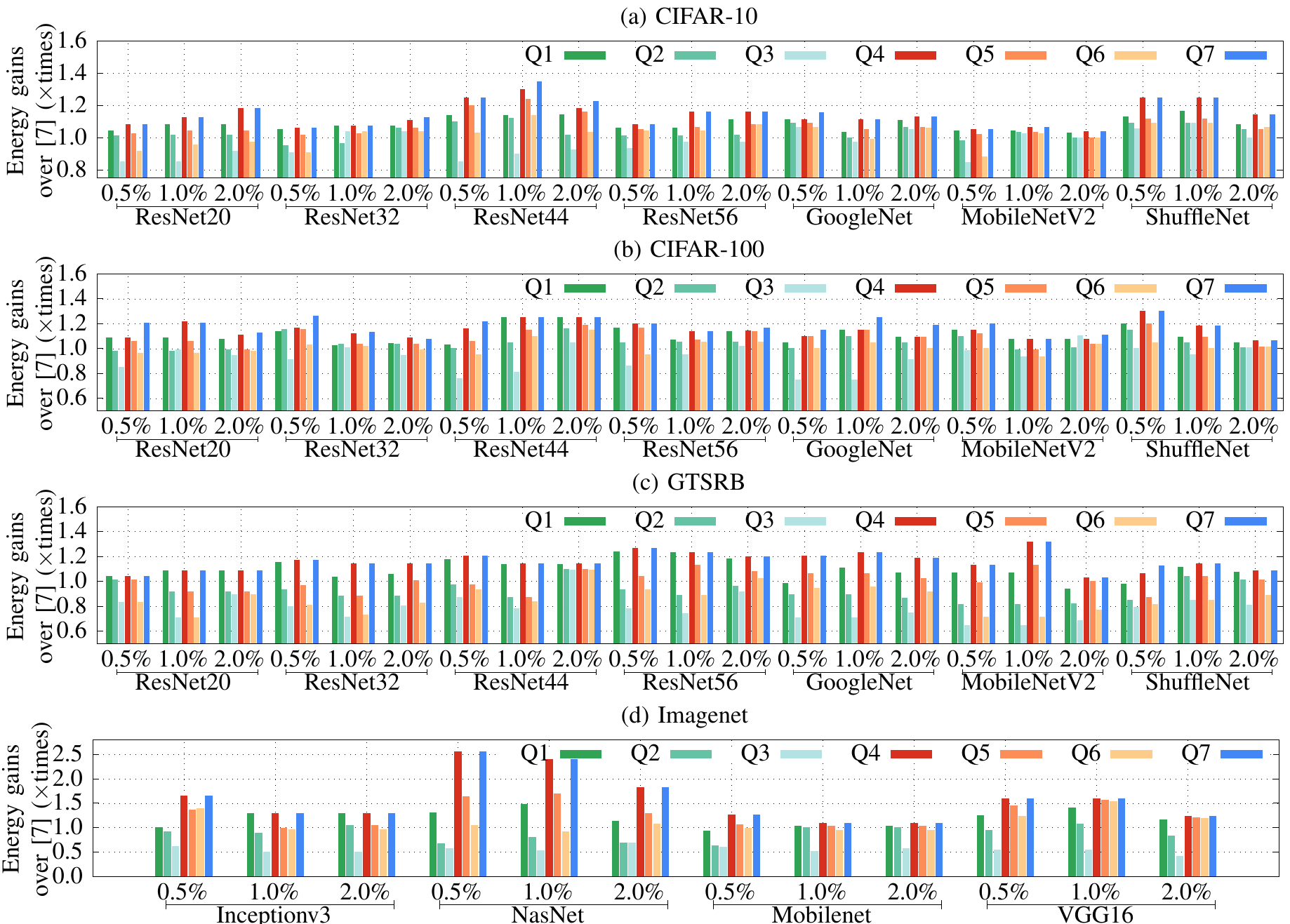}
\caption{Energy gains across all DNNs for (a) CIFAR-10, (b) CIFAR-100, (c) GTSRB, and (d) Imagenet over~\cite{tasoulas2020weight}. Q7 is the only constraint that~\cite{tasoulas2020weight} satisfies across all DNNs and datasets.}  
\label{fig:lvrm_energy}
\vspace{-15pt}
\end{figure*}

\begin{itemize}
	\item Generated weight-to-mode mapping solutions utilizing the LVRM multiplier by following their four-step methodology, which requires a single constraint regarding the average accuracy achieved over the dataset. Thus, we set $Accuracy_{thr,avg} = \{0.5\%, 1\%, 2\%\}$.
	\item Used the same approximate reconfigurable multiplier (LVRM) with our proposed mapping methodology, performed our proposed exploration, and
	generated different mappings per DNN, dataset and $Accuracy_{thr,avg}$ value.
\end{itemize}

In our motivation (Section~\ref{sec:motivation}) we mentioned that the mapping methodology used in \cite{tasoulas2020weight} is fairly aggressive and initially aims to map entire layers to the most aggressive approximate mode of the multiplier $M2$. To that end, the other approximate mode $M1$ is being underutilized. To further showcase this issue, Figure~\ref{fig:lvrm_distribution} depicts the utilization of the approximate modes across all layers of ResNet20 on CIFAR-10 between the mapping produced by~\cite{tasoulas2020weight} and our approach for Q7 and accuracy drop threshold of 1\%. It is evident that the mapping method in \cite{tasoulas2020weight} severely under-utilizes M1, leading to suboptimal solutions and inability to adequately benefit from this mode in terms of energy reduction. Contrary, our mapping achieved a more balanced utilization across the three multiplier modes, this time utilizing the $M1$ mode the most. Additionally, by assigning more operations to the $M1$ mode, the exact $M0$ mode is being utilized less; $20\%$ when compared to the $35\%$ $M0$ utilization by \cite{tasoulas2020weight}. \emph{This analysis validates our motivation that~\cite{tasoulas2020weight} loses efficient solutions due to biased decisions.}

First, we evaluated which queries out of the seven total shown in Table~\ref{tab:queries} were satisfied. \emph{Our method produced mapping solutions that satisfied all queries for all DNNs and datasets}. Table~\ref{tab:lvrm_queries} shows which queries the mapping methodology in~\cite{tasoulas2020weight} satisfied. 
Table~\ref{tab:lvrm_queries} does not quantitatively compares our method with LVRM. Rather, we demonstrate that although LVRM satisfies tight coarse constraints (Q7), it fails to satisfy finer ones (i.e. Q1-Q6).
LVRM~\cite{tasoulas2020weight} produces only one mapping that satisfies a general accuracy constraint threshold. Solutions towards more fine-grain optimizations are not supported, since the search way this method employs makes it infeasible to support more complex constraints.

Note that, the solution produced by LVRM is only one final mapping. We wanted to observe how demanding the more fine-grain accuracy requirements can be with a method that does not consider them at all.
We use the following notation:
\begin{inparaenum}
  \item the symbol \ding{55} indicates that the mapping did not satisfy the query for any of the three thresholds regarding average accuracy $Accuracy_{thr,avg} = \{0.5\%, 1\%, 2\%\}$;
  \item the symbol \ding{52} indicates that the mapping satisfied the query for all three thresholds; and
  \item in any other case, we list the threshold values under which the query was satisfied.
\end{inparaenum}
Table~\ref{tab:lvrm_queries} shows that the mapping methodology in~\cite{tasoulas2020weight} satisfies Q7 for all DNNs, datasets, and accuracy drop thresholds. This is expected as Q7 formally expresses the general required constraint that the average accuracy drop under approximation should be within specific values. However, it fails to satisfy Q2, Q3, and Q6 almost completely and especially for Imagenet, which is the most difficult examined dataset. These three queries require fine-grain exploration, which the method in~\cite{tasoulas2020weight} does not support, deeming it inappropriate for applications that require specific quality of service. Thus, it \emph{validates our motivation for supporting systematic fine-grain exploration}. 

Second, we compared the energy savings between the different methods. Particularly, Figure~\ref{fig:lvrm_energy} shows the energy gains of our mapping for each query over the solution found following the methodology in~\cite{tasoulas2020weight} for all DNNs, datasets, and average accuracy drop thresholds. 
For the CIFAR-10 dataset (Figure~\ref{fig:lvrm_energy}(a)), the gains of our method over LVRM~\cite{tasoulas2020weight} are overall lower than the ones achieved for the other datasets. This is attributed to the fact that CIFAR-10 is overall an ``easy'' dataset that comprises a number of 10 classes in total. Therefore, LVRM can achieve energy efficient mappings that leave little room for improvement; however the proposed method can exploit this tight margin to produce even better solutions in terms of energy gains. Respectively, the slightly more difficult GTSRB dataset (Figure~\ref{fig:lvrm_energy}(c)) that comprises 43 classes shows bigger gains over LVRM than the ones in CIFAR-10, and the even more difficult 100-class dataset CIFAR-100 (Figure~\ref{fig:lvrm_energy}(b)) shows the biggest energy gains overall. The hardest dataset evaluated in this work is the ImageNet which comprises 1000 classes, and shows the highest gains over LVRM overall. This is attributed to the fact that LVRM is mainly targetting mapping \emph{entire convolutional layers to the most aggressive multiplier mode}, which can lead to very pessimistic solutions. For instance, if mapping the two most error-resilient layers to the most aggressive multiplier mode violated the accuracy threshold, only the most error-resilient layer will be mapped to this mode entirely. For the rest of the layers, only ranges of weights are being examined for approximation mappings, which can provide very low final energy gains.

Based on the findings presented above, our \emph{methodology allows us to find mappings under multiple and fine-grain constraints, being also more energy-efficient than the state-of-art. Additionally, due to the formalization and the usage of PSTL, it is easy to create new queries and automate the search process (increased scalability), avoiding manual tuning}.

\vspace{-5pt}

\subsection{Comparison with a layer-oriented mapping methodology}

We compare our proposed methodology against ALWANN~\cite{mrazek2019alwann}, a layer-oriented method where a different static approximate multiplier from the EvoApprox library~\cite{mrazek2017evoapprox8b} is mapped to a DNN layer, leading to a different architecture for each distinct DNN. The authors consider a heterogeneous platform of set tiles, and employ a multi-objective genetic algorithm to map multipliers to each tile. Even though ALWANN does not propose a reconfigurable multiplier, the tile-based architecture allows the existence of multiple approximate multipliers at the same time. In our experimental evaluation, we consider the number of multipliers per tile to be 3. For the evaluation, we:
\begin{itemize}
	\item Generated layer-to-approximation mapping solutions following the procedure of ALWANN to select the multipliers. ALWANN also requires a constraint regarding the average accuracy achieved over the dataset, and thus we set $Accuracy_{thr,avg} = \{0.5\%, 1\%, 2\%\}$ for each DNN.
	\item We used the same approximate multipliers selected by ALWANN under our proposed mapping framework, performed the proposed exploration and generated different mappings per DNN, dataset and $Accuracy_{thr,avg}$ value.
\end{itemize}

\begin{table}[h]
\centering
\setlength{\tabcolsep}{3pt}
\caption{The queries ALWANN~\cite{mrazek2019alwann} satisfies for all datasets and DNNs}
\scriptsize
\label{tab:alwann_queries}
\resizebox{0.8\columnwidth}{!} 
{ 
\begin{tabular}{lccccccc}
\toprule
\textbf{CIFAR-10}    		& Q1      & Q2  & Q3  & Q4      & Q5  & Q6  & Q7            \\ \toprule \toprule
GoogLeNet   & \ding{55}       & \ding{55}   & \ding{55}  & 2\%       & 2\%   & 2\%   & \ding{52}     \\
MobileNetv2 & \ding{52} & \ding{55}   & \ding{55}  & \ding{52} & \ding{55}   & \ding{55}   & \ding{52} \\
ResNet20    & \ding{52} & \ding{55}   & \ding{55}  & \ding{52} & 1\%, 2\% & \ding{55}   & \ding{52} \\
ResNet32    & \ding{52} & 1\%, 2\% & 2\%  & \ding{52} & 1\%, 2\% & 1\%, 2\% & \ding{52} \\
ResNet44    & 1\%, 2\%     & 2\%   & \ding{55}  & \ding{52} & 2\%   & \ding{55}   & \ding{52} \\
ResNet56    & 2\%       & \ding{55}   & \ding{55}  & \ding{52} & 2\%   & \ding{55}   & \ding{52} \\
ShuffleNet  & \ding{52} & 2\%   & \ding{55}  & \ding{52} & 2\%   & \ding{55}   & \ding{52} \\ \bottomrule \\
\toprule
\textbf{CIFAR-100}   & Q1  & Q2 & Q3 & Q4  & Q5 & Q6 & Q7      \\ \toprule \toprule
GoogLeNet   & \ding{52} & 2\%   & \ding{55}  & \ding{52} & 2\%   & \ding{55}  & \ding{52} \\
MobileNetv2 & 1\%, 2\%     & 2\%   & \ding{55}  & 1\%, 2\%     & 2\%   & \ding{55}  & \ding{52} \\
ResNet20    & \ding{52} & 1\%, 2\% & \ding{55}  & \ding{52} & 1\%, 2\% & \ding{55}  & \ding{52} \\
ResNet32    & \ding{55}       & \ding{55}   & \ding{55}  & 1\%, 2\%     & \ding{55}   & \ding{55}  & \ding{52}     \\
ResNet44    & 1\%, 2\%     & 1\%, 2\% & \ding{55}  & 1\%, 2\%     & 1\%, 2\% & \ding{55}  & \ding{52} \\
ResNet56    & 2\%       & \ding{55}   & \ding{55}  & 2\%       & 2\%   & \ding{55}  & \ding{52}     \\
ShuffleNet  & 2\%       & \ding{55}   & \ding{55}  & \ding{52} & 1\%, 2\% & \ding{55}  & \ding{52} \\ \bottomrule \\
\toprule
\textbf{GTSRB}       & Q1  & Q2 & Q3 & Q4      & Q5 & Q6 & Q7      \\ \toprule \toprule
GoogLeNet   & \ding{55}       & \ding{55}  & \ding{55}  & \ding{55}       & \ding{55}  & \ding{55}  & \ding{52}       \\
MobileNetv2 & \ding{52} & \ding{55}  & \ding{55}  & \ding{52} & \ding{55}  & \ding{55}  & \ding{52} \\
ResNet20    & \ding{55}       & \ding{55}  & \ding{55}  & \ding{55}       & \ding{55}  & \ding{55}  & \ding{52} \\
ResNet32    & 2\%       & \ding{55}  & \ding{55}  & 2\%       & \ding{55}  & \ding{55}  & \ding{52} \\
ResNet44    & \ding{52} & \ding{55}  & \ding{55}  & \ding{52} & 2\%  & \ding{55}  & \ding{52} \\
ResNet56    & \ding{55}       & \ding{55}  & \ding{55}  & \ding{52} & \ding{55}  & \ding{55}  & \ding{52} \\
ShuffleNet  & \ding{52} & 2\%  & 2\%  & \ding{52} & 2\%  & 2\%  & \ding{52} \\ \bottomrule \\
\toprule
\textbf{ImageNet}    & Q1  & Q2 & Q3 & Q4  & Q5 & Q6 & Q7      \\ \toprule \toprule
Inceptionv3 & \ding{55}       & \ding{55}  & \ding{55}  & 1\%, 2\%     & \ding{55}  & \ding{55}  & \ding{52}     \\
Mobilenet   & \ding{52} & \ding{55}  & \ding{55}  & \ding{52} & \ding{55}  & \ding{55}  & \ding{52} \\
NasNet      & \ding{55}       & \ding{55}  & \ding{55}  & 2\%       & \ding{55}  & \ding{55}  & \ding{52}       \\
VGG16       & 1\%, 2\%     & \ding{55}  & \ding{55}  & 1\%, 2\%     & 2\%  & \ding{55}  & \ding{52} \\ \bottomrule
\end{tabular}
}
\vspace{-10pt}
\end{table}

First, we evaluated which queries out of the seven total shown in Table~\ref{tab:queries} were satisfied by ALWANN. As aforementioned, \emph{our method produced mapping solutions that satisfied all queries for all DNNs and datasets}. Table~\ref{tab:alwann_queries} shows which queries the ALWANN method satisfied. 
Again, we can see that ALWANN satisfies Q7 for all DNNs, datasets, and accuracy drop thresholds. This is expected because, as described previously, Q7 formally expresses the required constraint that the average accuracy drop should be within specific values. However, it fails to satisfy Q3 and Q6 almost completely and especially for Imagenet. 
In this case again, the solution produced by ALWANN is only one final mapping, and does not change across the queries. Solutions towards more fine-grain optimizations are not supported since the search way this method employs makes it infeasible to support more complex constraints.
At this point it is important to mention that since ALWANN follows a layer-based approach, the selected multipliers introduce smaller error with smaller energy reduction to satisfy the average accuracy thresholds. Consequently, ALWANN satisfies a lot of Q1 queries, while still having an impact on the achieved energy gains.

\begin{figure*}
\centering
\includegraphics{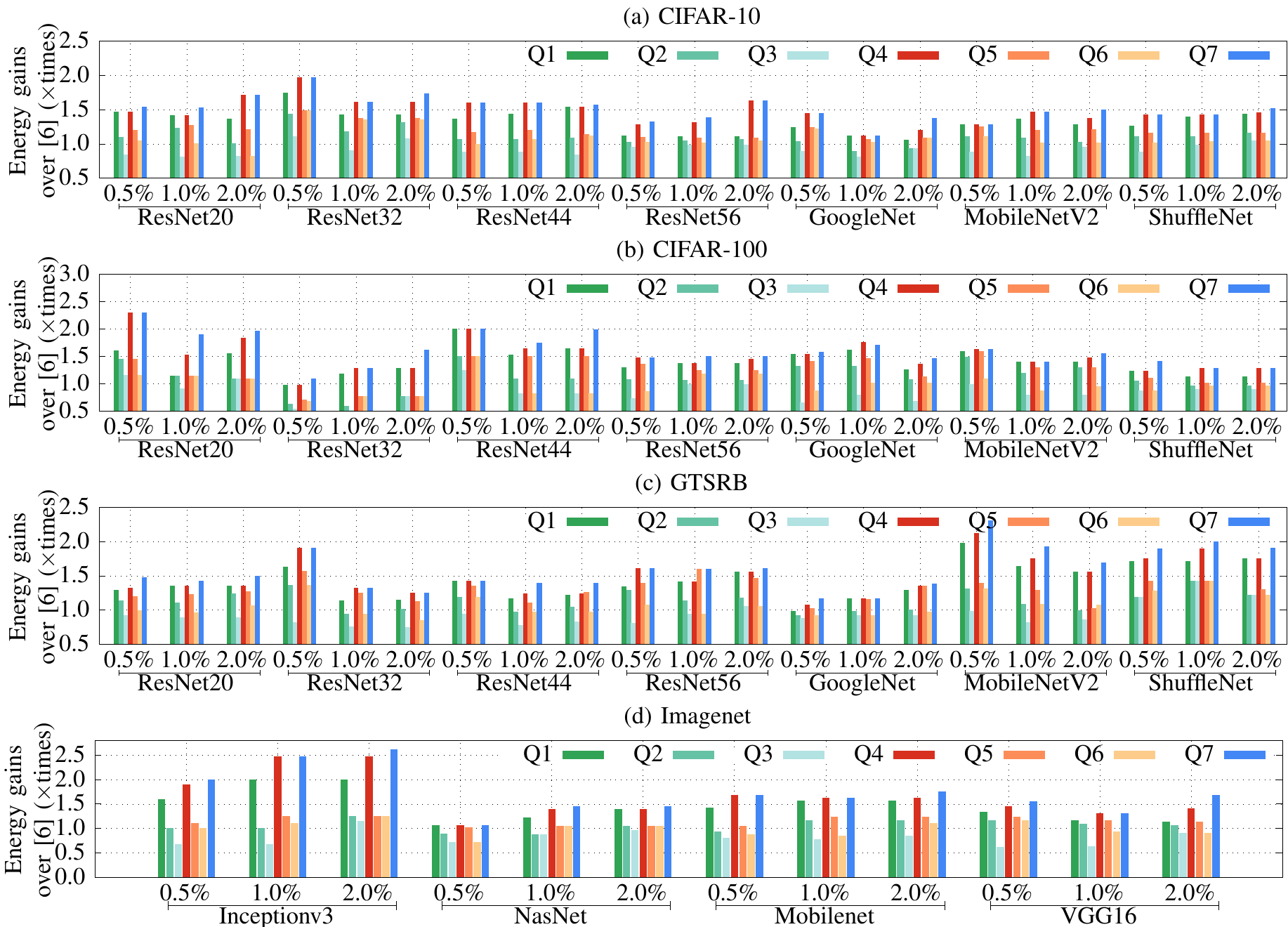}
\caption{Energy gains across all DNNs for (a) CIFAR-10, (b) CIFAR-100, (c) GTSRB, and (d) Imagenet over~\cite{mrazek2019alwann}. Q7 is the only constraint that~\cite{mrazek2019alwann} satisfies across all DNNs and datasets.}
\label{fig:alwann_energy}
\vspace{-15pt}
\end{figure*}

Figure~\ref{fig:alwann_energy} shows the energy gains of our proposed mapping methodology for each query over the solution found by ALWANN for all DNNs, datasets, and average accuracy drop thresholds. 
The proposed method produced mappings that achieved significantly higher gains in energy, since ALWANN is a layer-to-multiplier mapping approach while the proposed method follows a fine-grain weight mapping approach. Specifically for the ImageNet dataset (Figure~\ref{fig:alwann_energy} (d)), the multipliers selected by ALWANN were some of the least aggressive ones available~\cite{mrazek2017evoapprox8b} to satisfy the average accuracy constraints. 

\vspace{-5pt}

\subsection{Cost-effective Analysis}\label{sec:eval_exec_time}

All methods in our evaluation avoid retraining, therefore the execution cost lies in the amount of required inferences for each methodology. For instance, both methods in~\cite{tasoulas2020weight,spantidi2021positive} heavily depend on the examination of the error-tolerance for each distinct DNN layer (i.e., layer sensitivity). Consequently, these methods are guaranteed to run at least as many inferences as the amount of convolutional layers in a DNN. 

When compared to the ImageNet dataset, the CIFAR-10, CIFAR-100 and GTSRB datasets comprise images that are significantly smaller. They are resized to $32\times32$, while the images in ImageNet are rescaled to $224\times224$ ($299\times299$ for Inceptionv3). Therefore, inference is run significantly faster for the CIFAR-10, CIFAR-100 and GTSRB datasets. Additionally, since a larger image size leads to a larger amount of total multiplications that need to be computed on inference, for the ImageNet dataset step-based methods~\cite{tasoulas2020weight,spantidi2021positive} suffer in execution time. This fact, coupled with the vast amount of convolutional layers comprised by DNNs such as the Inceptionv3 or the NasNet, makes it infeasible to achieve a solution in an acceptable amount of time. For instance, for the VGG16 model on the Imagenet dataset, the work in~\cite{spantidi2021positive} required more than one day to finish a simple evaluation of approximate multipliers on a Xeon Gold 6230 processor running at 2.10GHz, deeming it not scalable for larger datasets, thus we did not include it in our evaluation. 

To ensure the proposed framework will always produce mapping solutions in acceptable timeframes, we pre-define a set number of optimization iterations. Specifically, we define the number of iterations for the optimization to be 50 for the CIFAR-10, CIFAR-100 and GTSRB datasets, and 100 for the ImageNet dataset. 
The proposed framework found the solution for a single query 45\% faster on average compared to the complete 4-step exploration of~\cite{tasoulas2020weight}. This speed boost compared to~\cite{tasoulas2020weight} allowed us to include more queries in our evaluation, providing greater insight on the impact of fine-grain accuracy constraints and approximate multipliers on the energy gains of a DNN accelerator. The inclusion of ERGMC and robustness calculation in this work inflict negligible time overhead.

\section{Conclusion}

In this work, we present a unified framework that uses formal properties 
of approximate DNN accelerators that support reconfigurable approximate multiplications
, thus  enabling fine-grain optimizations. We utilize specification formalisms to express DNN properties and employ stochastic optimization to find approximate mappings that satisfy accuracy thresholds. 
We conducted an in-depth evaluation across multiple DNNs, datasets and accuracy requirements, to obtain a better grasp of how strict constraints can still yield energy savings.
When compared to other mapping methodologies, on the same multipliers, our framework achieves even more than $\times2$ the energy gains without violating the defined accuracy thresholds. 

\vspace{-5pt}

\bibliographystyle{IEEEtran} 
\bibliography{bibliography}
\vspace{12pt}

\end{document}